%% file: main.tex
\documentclass[letterpaper,10pt,conference]{IEEEtran}
\IEEEoverridecommandlockouts

\usepackage[T1]{fontenc}
\usepackage[utf8]{inputenc}
\usepackage{cite}
\usepackage{graphicx}

\usepackage{amsmath,amssymb,bm,enumitem}
\usepackage{booktabs,tabularx,array,multirow}
\usepackage{xcolor}
\usepackage[normalem]{ulem}
\usepackage{url}

\usepackage[
  letterpaper,
  left=0.75in,
  right=0.75in,
  top=1in,
  bottom=0.75in,
  headheight=0pt,
  headsep=0pt,
  footskip=0pt
]{geometry}

\usepackage[hidelinks]{hyperref}
\newcommand{\paragraphc}[1]{\vspace{0.2em}\noindent\textbf{#1}}

\newif\ifdraftnotes
\draftnotestrue
\definecolor{wenhan}{HTML}{2E8B57}

\newcommand{\method}{\textsc{HDH}}
\newcommand{\pifive}{\ensuremath{\pi_{0.5}}}
\newcommand{\pending}{\ifmmode\text{\textemdash}\else\textemdash\fi}

\makeatletter
\newcommand{\IEEEfullwidthfigure}[3]{%
  \refstepcounter{figure}%
  \centering
  #1\par
  \vspace{2pt}%
  \@makecaption{\fnum@figure}{#2}%
  \label{#3}%
}
\makeatother

\title{\LARGE\bfseries Learning Low-Latency Action Verification for Frozen Vision-Language-Action Models}

\author{Wen-Han Hsieh$^{*,1}$, Ken Jiankun Zheng$^{*,1}$, Chia-Chien Chen$^{*,2}$, David M. Chan$^{1}$, Trevor Darrell$^{1}$%
\thanks{$^{*}$Equal contribution.}%
\thanks{$^{1}$University of California, Berkeley}%
\thanks{$^{2}$National Tsing Hua University}%
\thanks{Project website: \url{https://wen-hanhsieh.github.io/hdh/}}%
}

\IEEEaftertitletext{%
  \IEEEfullwidthfigure
    {\includegraphics[width=\linewidth]{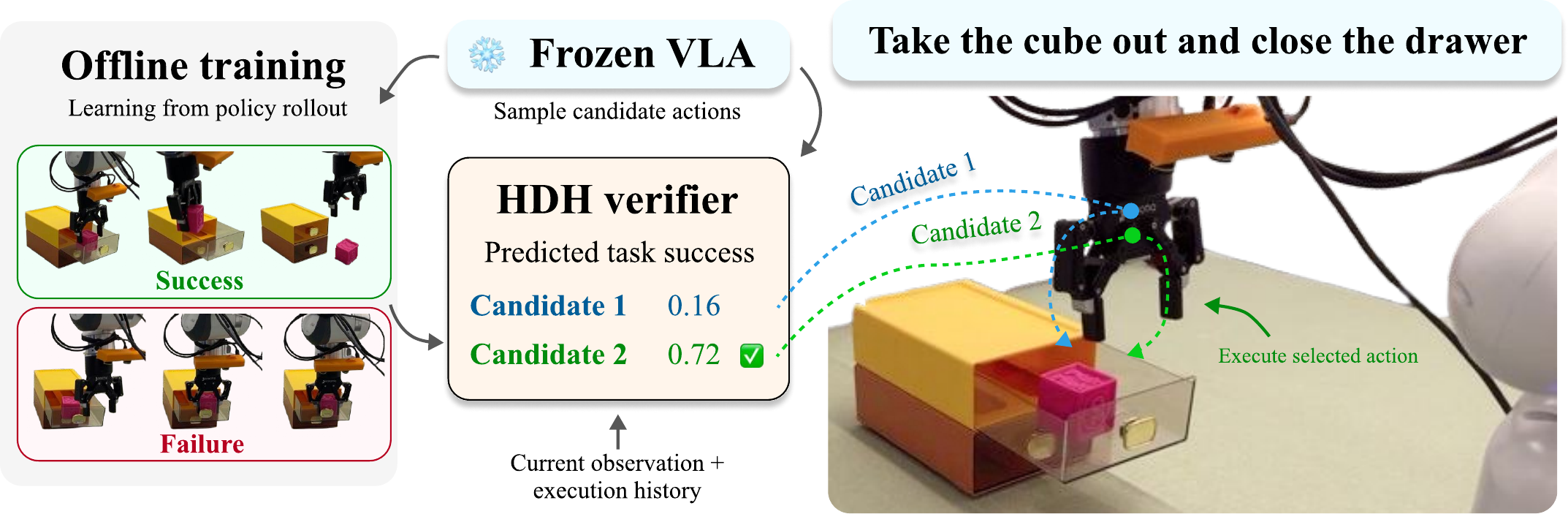}}
    {We propose \method{}, a lightweight framework for improving action selection in frozen vision-language-action models. We first collect autonomous policy rollouts and use their terminal success and failure outcomes to train a compact verifier (left). At each policy query, the verifier uses the current observation and execution history to score sampled action chunks (center), and the robot executes the action of the highest-scoring candidate (right).}
    {fig:vsa-overview}%
  \vspace{6pt}%
}

\begin{document}
\maketitle
\thispagestyle{empty}
\pagestyle{empty}

\begin{abstract}
Vision-language-action (VLA) models have made significant progress in general-purpose robotic manipulation. However, sampled action sequences can be ineffective, while evaluating alternatives with computationally expensive verifiers adds latency at each policy query. In this work, we propose \method{}, a lightweight framework that uses hindsight from a frozen VLA's own deployments to improve action verification. The framework has three stages. First, we deploy the policy and collect its autonomous rollouts. We then train a compact action verifier to assign each action chunk a predicted task-success score. Finally, the frozen VLA generates action candidates, and the verifier selects the candidate with the highest score for execution. Across four LIBERO-10 tasks, \method{} achieves an average success rate of 82.3\% and on three real-world manipulation tasks, \method{} improves average success over the baseline policy by 22.6\%.
\end{abstract}


\input{sections/introduction}
\input{sections/related_work}

\begin{figure*}[!t]
  \centering
  \includegraphics[width=\linewidth]{\detokenize{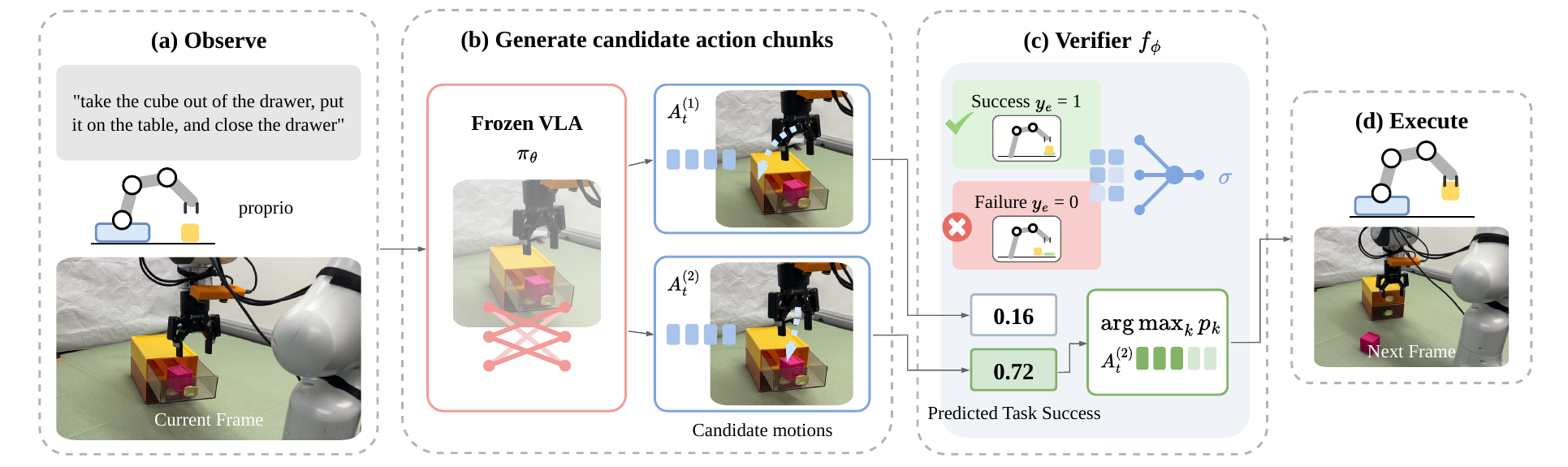}}
    \caption{\textbf{Overview of \method{}} (a) Given the task instruction, current image observations, and proprioception, (b) we sample candidate action chunks from a frozen VLA. (c) We assign each candidate a predicted task-success score using a compact verifier trained on terminal rollout outcomes. (d) The robot executes the first $H$ actions of the candidate with the highest predicted task-success score.}
  \label{fig:vsa-action-selection}
\end{figure*}

\input{sections/method}
\input{sections/experiments}

\section{Conclusion and Future Work}


We present \method{}, a lightweight action verifier that improves action selection without retraining the VLA. Our framework consists of three stages. First, we collect autonomous rollouts from the frozen policy. We then use their terminal outcomes to train a compact verifier that assigns a predicted task-success score to each candidate action chunk. Finally, we sample action candidates from the frozen VLA and execute the highest-scoring candidate. Average task success increases from 78.8\% to 82.3\% in simulation and from 44.7\% to 67.3\% in the real-world. In simulation, median rollout time per query increases by only 5.4\% over the baseline and remains substantially below that of the evaluated VLM verifiers. These results show that learned action verification can improve a frozen VLA with a modest increase in overall query time.

\paragraphc{Limitations and Future Work.}
Our verifiers are trained separately for each task; learning a shared verifier could reduce data collection and training requirements across tasks. Terminal outcomes also provide coarse supervision, motivating intermediate feedback to better distinguish useful action choices. Our timing measurements include simulation overhead. Although the verifier operates smoothly in our real-world experiments, future work should measure the real-world latency breakdown. 
Our experiments utilize a candidate set of two actions, leaving open how verifier performance and computational cost would scale with larger candidate sets.
Finally, our evaluation covers a limited number of tasks and robot settings, so broader experiments to more thoroughly assess transfer across tasks, embodiments, and environments would be a beneficial direction for future work.

\section*{Acknowledgments}
We thank Elvis Hsieh and Steve Chen for their invaluable feedback during the discussions. Authors, as part of their affiliation with UC Berkeley, were supported in part by the National Science Foundation, US Department of War, and/or the Berkeley Artificial Intelligence Research (BAIR) industrial alliance program, as well as an invaluable gift from Arcee AI. This research was also developed with funding from the Defense Advanced Research Projects Agency (DARPA) under Contract No. HR0011-25-3-0133. The views, opinions and/or findings expressed are those of the authors and should not be interpreted as representing the official views or policies of any sponsor, the Department of War, or the U.S. Government.


\bibliographystyle{IEEEtran}
\bibliography{references}

\clearpage
\input{sections/appendix}

\end{document}

%% file: sections/introduction.tex
\section{Introduction}
\label{sec:introduction}

Vision-language-action (VLA) models have achieved state-of-the-art performance on general-purpose manipulation, mapping visual observations and language instructions into robot actions~\cite{rt2,openvla,pi0,pi05}. In deployment, however, the common practice is to sample a single action from the VLA, and execute it on the robot. Existing deployments, thus, do not exploit a fundamental property of generative policies, where the same observation and instruction can yield a full \textit{distribution} of candidates with substantial variation in execution quality. To exploit this distributional generation capability, a simple test-time solution is to sample multiple candidate action chunks and select the candidate with the highest estimated quality. 

The question of how to select that action is, however, an open problem. Existing work has considered several heuristic selection strategies, including random selection and candidate evaluation using large vision-language models~\cite{vgps,robomonkey,forewarn,rover}. However, we believe that instead of relying on very large external models, the policy's own deployment experience can provide explicit supervision for this choice. We therefore investigate whether terminal success and failure provide sufficient supervision for candidate ranking and whether a compact verifier can achieve selection performance comparable to that of substantially larger verifiers.

To address these questions, we introduce Hindsight Decision Model with History (\method{}), a lightweight verifier trained on terminal outcomes from autonomous policy rollouts, without expert comparisons, synthetic preferences, or predicted futures. Each training example consists of the current observation, the executed action prefix, and the episode's terminal outcome. To provide context about task progress, the verifier additionally receives recent observations and executed actions. At deployment, \method{} assigns a predicted eventual task-success score to each candidate sampled from the frozen VLA, and the robot executes the highest-scoring prefix.

We evaluate \method{} in simulation and in the real-world in four dimensions. First, does a verifier trained only from terminal outcomes improve the closed-loop success of a frozen VLA? Second, does it produce candidate rankings that improve selection performance, beyond randomly selecting from the same set? Third, does history-conditioned scoring improve candidate discrimination by revealing task progress? Finally, how does such a compact verifier compare with verifiers based on large vision-language models on task success and inference latency?

We find that hindsight supervision alone substantially improves closed-loop manipulation. Across four LIBERO-10 tasks, \method{} achieves an average success rate of 82.3\%, outperforming both the frozen policy and random selection from the same candidate set. On three real-world manipulation tasks, it raises average success from 44.7\% to 67.3\% within a real-time control budget. Furthermore, history-conditioned scoring improves candidate ranking, and the compact verifier achieves selection performance comparable to that of much larger verifiers with substantially lower inference cost.

Our contributions are thus threefold:
\begin{itemize}
    \item \textbf{Problem setting.} We frame VLA deployment as real-time candidate selection supervised only by the frozen policy's own rollout outcomes.
    \item \textbf{Verifier and selection.} We introduce a lightweight, history-conditioned verifier for ranking native VLA candidates without updating the policy.
    \item \textbf{Evaluation.} We show in both simulation and real-world experiments that hindsight-supervised selection improves closed-loop manipulation with substantially lower inference cost than large verifiers.
\end{itemize}

%% file: sections/related_work.tex
\section{Related Work}

\subsection{Generative Robot Policies and Test-Time Sampling}

Large-scale robot datasets have enabled policies that can generalize across diverse tasks, environments, and embodiments~\cite{oxe,bridge,droid}. Recent vision-language-action models further combine visual and language representations with expressive action generators~\cite{rt2,openvla,octo,cogact}. Diffusion Policy models multimodal action distributions through iterative denoising~\cite{diffusionpolicy}, while the $\pi_0$ family uses flow matching to generate continuous action chunks~\cite{pi0,pi05}. Because these policies represent distributions over actions, repeated sampling can produce different behaviors for the same observation and instruction. Similar test-time sampling strategies have improved language-model inference by generating multiple candidate responses and selecting among them~\cite{cobbe,monkeys,snell}. In robot control, however, additional samples are useful only when the system can identify which candidate is most likely to advance the task. Our work keeps a fine-tuned \pifive{} policy fixed and learns to select among its sampled action chunks using outcomes from prior policy rollouts.

\subsection{Action Verification and Policy Steering}

Several approaches improve robot behavior by evaluating multiple candidate actions before execution. V-GPS reranks actions using a value function learned through offline reinforcement learning, without modifying the original policy~\cite{vgps}. IDQL similarly combines expressive diffusion behavior models with critic-based policy extraction~\cite{idql}. RoboMonkey generates synthetic action preferences from expert demonstrations and trains a vision-language model to select among policy candidates~\cite{robomonkey}, while RoVer and CoVer-VLA learn evaluators that assess candidate quality and consistency with the current observation and instruction~\cite{rover,cover}. Other methods explicitly estimate the consequences of an action: FOREWARN predicts future latent states with a world model and evaluates them using a vision-language model~\cite{forewarn}, whereas inference-time policy steering can incorporate human guidance into generative sampling~\cite{itps}. DSRL instead learns to steer the latent noise supplied to a frozen generative policy~\cite{dsrl}. These methods establish that inference-time evaluation can improve pre-trained policies, but differ in how candidates are produced, how supervision is obtained, and how much computation is required. In contrast, \method{} scores the frozen VLA's native action candidates with a compact verifier trained directly from terminal outcomes of its own rollouts.

\subsection{Learning from Policy Rollouts}

Policy rollouts can also improve behavior after initial policy training. Autonomous improvement systems collect additional interactions beyond expert demonstrations~\cite{autonomous}; RECAP uses this experience to estimate action values and fine-tune the VLA toward more successful behavior~\cite{recap}. Policy-optimization methods adapt diffusion policies using reward feedback~\cite{ddpo,dppo}, while GRAPE, SafeVLA, and RL Token provide alternative interfaces for preference alignment, constrained adaptation, or compact actor--critic refinement~\cite{grape,safevla,rltoken}. These approaches primarily use policy rollouts to update or steer the policy. A related line of work uses execution data for failure detection rather than policy improvement. Sentinel monitors temporal action consistency and semantic task progress~\cite{sentinel}, and SAFE detects failures from internal VLA representations~\cite{safe}. Such monitors can predict when execution is likely to fail, but do not necessarily determine which alternative action should be executed. In contrast, \method{} uses terminal rollout outcomes to train a verifier that ranks the frozen policy's candidate actions and directly determines which action is executed.

%% file: sections/method.tex
\section{Learning Action Verification from Policy Rollouts}
\label{sec:method}

An overview of our method is shown in Fig.~\ref{fig:vsa-action-selection}. First, we collect rollouts from a frozen VLA and use their terminal outcomes to supervise the recorded actions. We train a \emph{Hindsight Decision Model} (HDM) to score candidate actions using the current observation and proprioceptive state. Our full model, the \emph{Hindsight Decision Model with History} (HDH), additionally incorporates recent observations and executed actions. During closed-loop control, the frozen VLA samples multiple action candidates, and the verifier selects the candidate with the highest predicted task-success score.

\subsection{Action Chunk Success Estimation}
Let $o_t$ contain the camera observations and proprioception at decision $t$, and let $\ell$ denote the instruction. A chunk $A_t\in\mathbb{R}^{L\times d_a}$ is generated by sampling from the frozen policy $\pi_\theta$ using these inputs and flow-matching noise (the frozen policy's action-generation noise). The robot executes only $U_t=A_t[0:H]$, after which the system samples a new plan using the resulting observation. The verifier context $x_t$ is defined to include the current observation and, for the history-conditioned configuration, information from prior observations and executed actions recorded before decision $t$.

The quantity of interest is the probability of eventual task success after $U_t$ is executed and subsequent actions are generated by the baseline policy:
\begin{equation}
Q^{\pi_\theta}(x_t,U_t)=\Pr(y_e=1\mid x_t,U_t;\pi_\theta\text{ thereafter}),
\label{eq:value}
\end{equation}
where $y_e \in \{0,1\}$ denotes terminal episode success. We approximate this quantity as $Q_\phi(x_t,U_t)=1-\sigma(f_\phi(x_t,U_t))$, where $f_\phi$ is a failure logit. Thus, $Q_\phi$ represents the predicted eventual task-success score learned from observed rollout outcomes.

\subsection{Learning from Terminal Outcomes}
Each retained example consists of a decision context, its actually executed prefix, and the terminal outcome of that episode. With failure label $y_e^{\mathrm{fail}}=1-y_e$, the simulation objective is
\begin{equation}
\mathcal{L}_{\mathrm{MC}}=\frac{1}{N_q}\sum_{(e,t)}
\operatorname{BCEWithLogits}\!\left(f_\phi(x_t,U_t),1-y_e\right).
\label{eq:outcome}
\end{equation}
The same episode return is assigned to all retained queries from that episode. The dataset is partitioned at the episode level, rather than by randomly assigning queries from the same trajectory across splits.

Under the current preprocessing convention, queries whose recorded executed length is below $H$ are excluded. This convention prevents supervision from being applied to action rows that were not executed, especially at terminal queries.

\subsection{Action Verifier Architecture}
The core attention-based scoring path is shown in Fig.~\ref{fig:vsa-action-selection}. In HDM, frozen backbone features are extracted from two image views (early-layer outputs of the backbone's SigLIP image pathway)~\cite{siglip,pi05}. After region pooling, the representation contains an $8\times8$ grid for each view, yielding 128 visual tokens of feature dimension 2048. These features are projected to width 128, with spatial and view embeddings. An eight-dimensional proprioceptive state is projected into one additional token. A separate verifier is used for each task, with a constant task token. The scorer input does not include language-prompt tokens, although the policy remains instruction-conditioned.

The scorer input includes the $H\times7$ executed prefix, typically $H=8$ from a length-$10$ predicted chunk. Action tokens are formed using a linear action projection and learned temporal embeddings. The architecture applies four-head action self-attention, followed by cross-attention from the action tokens to the observation tokens~\cite{attention}, with residual connections, normalization, and a feed-forward network. The failure logit is computed by mean pooling over the action tokens followed by an MLP of widths $128\rightarrow64\rightarrow1$. We convert this logit into the task-success score $Q_\phi(x_t,U_t)=1-\sigma(f_\phi(x_t,U_t))$ for candidate selection. The reported current-context implementation has approximately 481K trainable parameters. The policy parameters remain frozen.


In the history-conditioned variant (HDH), a causal history token (a single token summarizing only past observations and executed actions) is computed from prior observations, proprioception, and executed actions using a small GRU-based encoder and added to the representation. The history state is reset between episodes and is computed without future observations, future actions, or terminal outcomes at inference time. Parameters for both HDM and HDH are optimized using the Monte Carlo objective in Eq.~\ref{eq:outcome}. The contribution of history is evaluated through a matched comparison of HDM and HDH.

\subsection{Candidate Sampling and Action Selection}
At each decision, $K=2$ candidates are generated by independently sampling from the frozen policy using independently seeded flow-matching noise (a separate random seed per candidate), with identical current observations and instructions. For candidate prefixes $U_t^{(k)}=A_t^{(k)}[0:H]$, selection is
\begin{equation}
k^*=\arg\max_{k\in\{1,\ldots,K\}} Q_\phi\!\left(x_t,U_t^{(k)}\right).
\end{equation}
The robot executes the prefix with the highest predicted task-success score without modification. After the next observation is obtained, the selection procedure is applied again. The native $K=1$ baseline produces stochastic outputs because each query uses a separately sampled noise realization. With independent identically distributed native candidates, the marginal action distribution under uniform random selection of one member of the $K=2$ set is identical to that of $K=1$. The random-selection condition therefore provides a control comparison for the candidate-generation and selection interface. By construction, the marginal candidate distribution under random selection is identical to that of the $K=1$ baseline.

The implementation can reuse observation encodings and the frozen prefix's cached attention keys and values (KV cache) across candidates within a query, batch flow trajectories, and reuse observation projections that are unchanged across candidates while batching verifier scores. When the same random seed is reused, candidate-generation results can be compared only for identical inputs. Once different actions are selected, trajectories may diverge, and later observations and candidate chunks can differ.

%% file: sections/experiments.tex
\begin{figure*}[t]
\centering
\includegraphics[width=\linewidth]{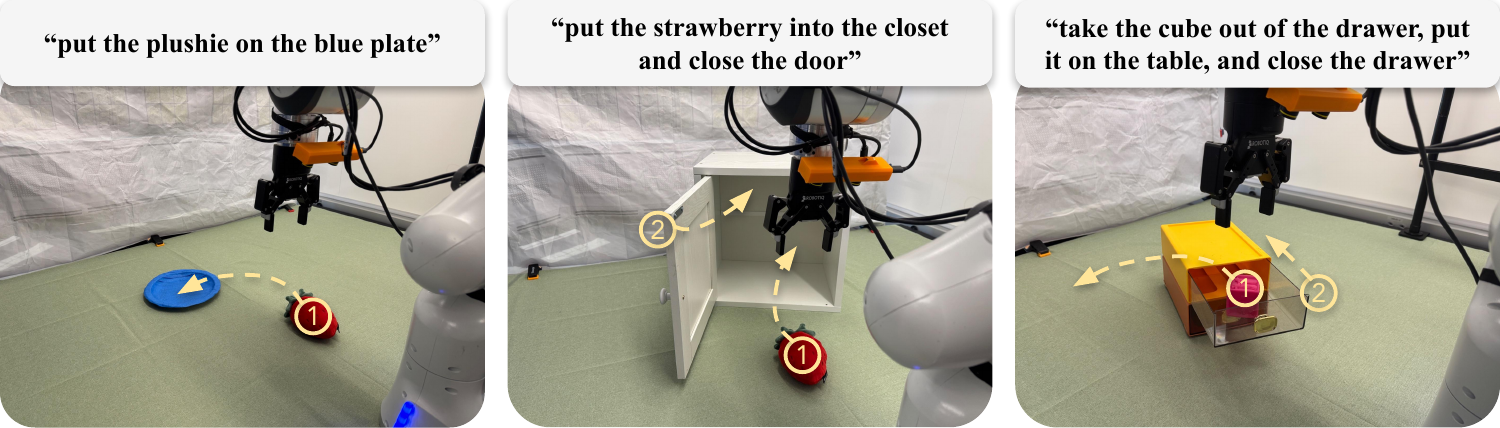}
\caption{\textbf{Real-world manipulation tasks.} We evaluate placing a plushie on a plate, placing a strawberry in a closet and closing the door, and removing a cube from a drawer and closing the drawer. These tasks span object placement and multi-stage manipulation.}
\label{fig:real_tasks}
\end{figure*}

\begin{figure}[t]
\centering
\includegraphics[width=\linewidth]{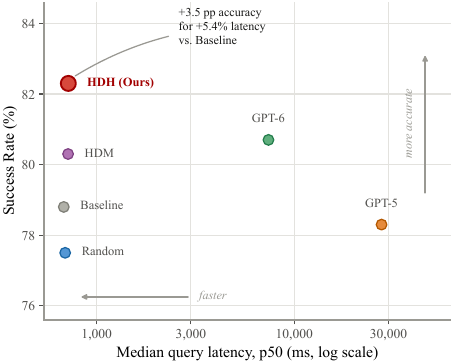}
\caption{\textbf{Accuracy--latency trade-off on LIBERO.} \method{} achieves the highest average success rate, improving the baseline by 3.5\% with only 5.4\% additional median query latency. Latency is shown on a logarithmic scale.}
\label{fig:libero_latency_tradeoff}
\end{figure}

\begin{figure*}[t]
\centering
\includegraphics[width=\linewidth]{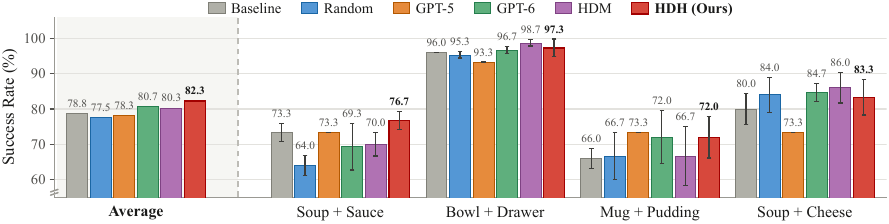}
\caption{\textbf{LIBERO simulation results.} Success rates across four manipulation tasks and their unweighted average. Bars show the mean, and error bars indicate standard deviation across runs.}
\label{fig:libero_success_rates}
\end{figure*}

\begin{figure}[t]
\centering
\includegraphics[width=\linewidth]{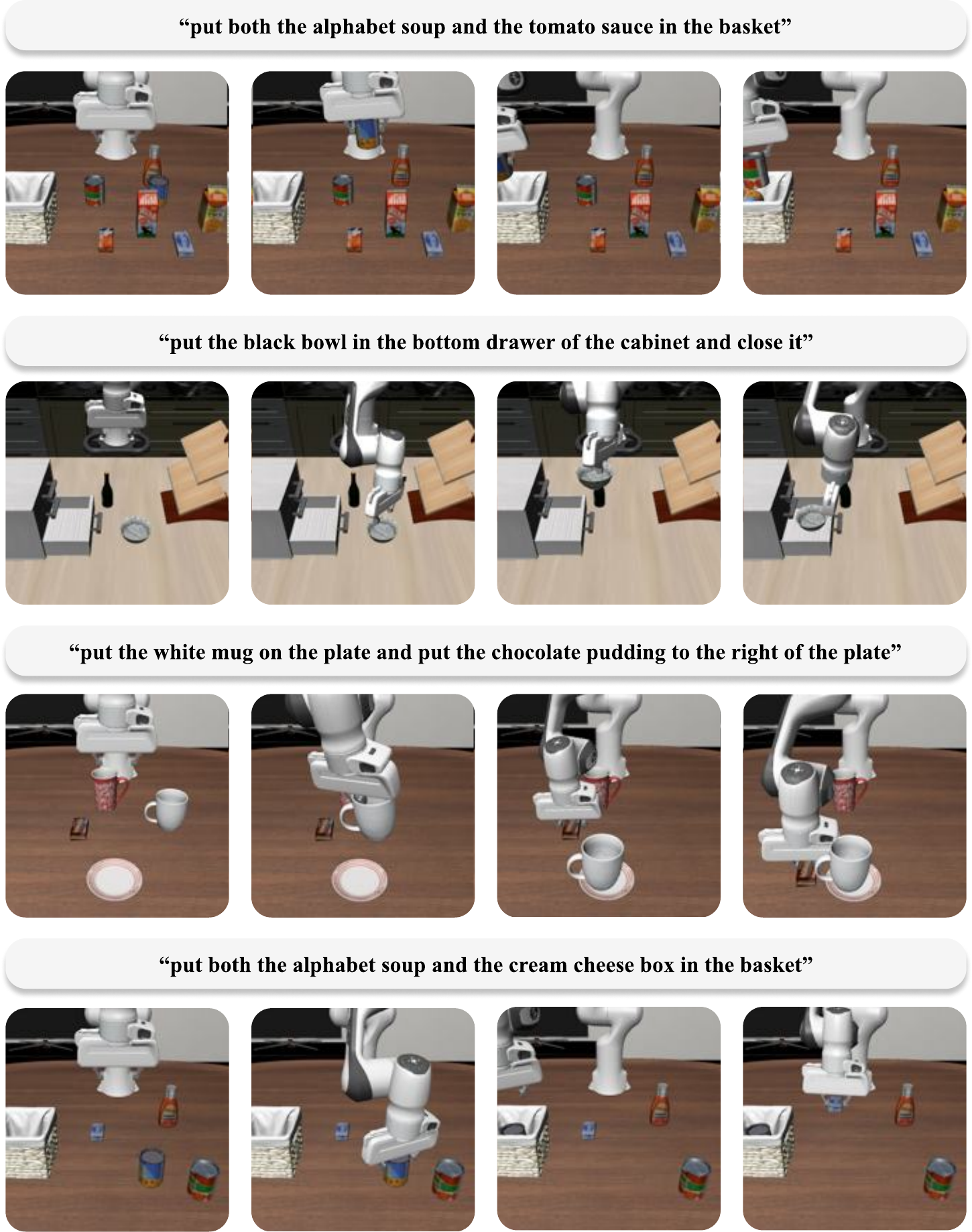}
\caption{\textbf{LIBERO-10 evaluation tasks.} From top to bottom: placing alphabet soup and tomato sauce into a basket; placing a black bowl into the bottom drawer and closing it; placing a white mug on a plate and chocolate pudding to its right; and placing alphabet soup and cream cheese into a basket. Each row shows successive stages of task execution from left to right.}
\label{fig:simulation_tasks}
\end{figure}

\section{Experiments \& Results}
\label{sec:experiments}

In this section, we compare \method{} with the baseline policy and alternative action verifiers in simulation, then evaluate it against the baseline policy on real-world manipulation tasks. Specifically, we study: (1) whether learning from terminal rollout outcomes improves task success beyond the frozen policy and random candidate selection; (2) whether execution history improves action selection; (3) how a compact verifier compares with large vision-language model verifiers in task success and computational cost; and (4) whether the approach improves task completion on a physical robot.

\subsection{Experimental Setup}
\label{subsec:setup}

\paragraphc{Simulation Tasks.}
We report results on four multi-stage manipulation tasks from LIBERO-10~\cite{libero}: placing alphabet soup and tomato sauce into a basket; placing a black bowl into the bottom drawer and closing it; placing a white mug on a plate and chocolate pudding to its right; and placing alphabet soup and cream cheese into a basket. 

\paragraphc{Baseline Policy.}
We initialize \pifive{} from its pretrained checkpoint and jointly fine-tune it on 50 public LIBERO demonstrations across five training tasks, with ten demonstrations per task. Fine-tuning uses LoRA on the Gemma-2B backbone and 300M action expert for 5,000 steps with a global batch size of 32. All six methods use the same frozen policy, which predicts ten-step action chunks and executes eight steps before the next query.

\paragraphc{Verifier Training.}
We collect 100 rollouts per task using the frozen baseline policy with one sampled action chunk per query. Terminal success or failure provides Monte Carlo supervision for each retained observation and executed action sequence. Each task's dataset is divided into 80 training and 20 validation episodes. We train separate HDM and HDH verifiers for each task using three training seeds matched across the two variants. HDM uses the current observation and action sequence, while HDH additionally incorporates observations and executed actions from up to three preceding completed queries.

\paragraphc{Compared Methods.}
We compare six methods using the same frozen VLA. Baseline executes a single sampled action chunk. The remaining methods select between two sampled candidates: Random selects uniformly, GPT-5 and GPT-6 use VLM-based verification, and HDM and HDH use the learned verifiers described in Section~\ref{sec:method}.

Both VLM verifiers receive the task instruction, current external and wrist-camera images, robot state, and the first eight actions of each candidate chunk, together with a history of external-camera images, end-effector positions, gripper states, and executed actions from up to three previous policy queries. We use \texttt{gpt-5} and \texttt{gpt-6-astra} with the same prompt format and rules for interpreting their outputs. Failed requests or invalid outputs are retried up to twice. If all attempts fail, the robot executes the first sampled candidate, and the episode remains in the evaluation.

\paragraphc{Evaluation Protocol.}
We evaluate each method on 50 held-out initial states per task over three runs with different random seeds, totaling 600 episodes across four tasks. GPT-5 uses 15 states per task in a single run (60 episodes) due to slower inference. Initial states are shared across methods and runs and are disjoint from verifier-training initial states. For HDM and HDH, each run also uses a separately trained verifier. We report mean success and standard deviation across runs, except for GPT-5, and the unweighted average success across tasks.

\subsection{Does Learned Action Selection Improve Task Success?}
\label{subsec:simulation-results}

We compare HDH with the frozen baseline policy, which executes one sampled action chunk, and Random, which selects uniformly between two candidates generated using the same sampling procedure as HDH.

As shown in Figure~\ref{fig:libero_success_rates}, HDH achieves 82.3\% average success across four tasks, compared with 78.8\% for the baseline and 77.5\% for Random. HDH also achieves higher observed success than the baseline on every evaluated task.

The advantage over Random suggests that the gains come from learned candidate ranking rather than additional sampling alone. The VLA remains unchanged: the verifier changes which sampled action is executed, not how candidates are generated.

\subsection{Does Execution History Improve Action Selection?}
\label{subsec:history-results}

We compare HDH with HDM to assess the contribution of execution history. Both variants use the same frozen policy, rollout dataset, candidate sampling procedure, and Monte Carlo objective, with matched training seeds. HDM scores candidates using the current observation, whereas HDH additionally incorporates observations and executed actions from up to three previous completed queries. As shown in Figure~\ref{fig:libero_success_rates}, adding history increases average success from 80.3\% to 82.3\%, with the largest gains on Soup + Sauce (+6.7\%) and Mug + Pudding (+5.3\%). Both tasks require coordinating successive interactions with multiple objects. Recent observations and executed actions may help the verifier distinguish completed subgoals from unsuccessful attempts, favoring candidates that advance the task rather than repeat ineffective actions. These gains suggest that temporal context can support action selection beyond the current observation alone.


\subsection{How Does the Compact Verifier Compare with VLM Verifiers?}
\label{subsec:computational-cost}

We compare HDH with GPT-5 and GPT-6 in terms of task success and computational cost. All three methods select between two action candidates generated by the same frozen policy. HDH achieves the highest observed average success among the evaluated methods, as shown in Figure~\ref{fig:libero_success_rates}. GPT-5 is evaluated on fewer trials, and both VLM verifiers receive execution history without wrist images or end-effector orientation.

\input{tables/inference_time}

Figure~\ref{fig:libero_latency_tradeoff} summarizes the accuracy--latency trade-off, while Table~\ref{tab:inference_time} reports the corresponding p50 and p95 query times. Compared with the baseline, HDH improves average success by 3.5 percentage points while increasing median query latency by only 5.4\%. In contrast, GPT-5 and GPT-6 require substantially longer query times without achieving higher average success than HDH.

Policy inference runs on a single NVIDIA H100 80GB GPU. We measure end-to-end query time as episode duration divided by the number of policy queries, including simulator overhead and, for the VLM verifiers, network and service waiting. These results show that the compact verifier improves action selection while keeping query latency close to that of the frozen baseline.


\input{tables/realworld}

\subsection{Does Action Selection Improve Real-World Task Success?}
\label{subsec:realworld-results}

\paragraphc{Setup.}
We evaluate on a Franka Panda using the DROID interface~\cite{droid} across the three tasks shown in Figure~\ref{fig:real_tasks}: placing a plushie on a plate, removing a cube from a drawer and closing it, and placing a strawberry into a closet and closing the door. The policy receives external and wrist RGB images and proprioception, and controls joint velocities and gripper position at 15 Hz. We fine-tune \pifive{}-DROID using LoRA on 80 demonstrations across four training tasks for 5,000 steps. The verifier dataset contains 150 retained rollouts, partitioned into 80\% training and 20\% validation episodes.

We use the same HDH architecture and Monte Carlo objective as in simulation, incorporating up to three previous completed queries. The selector compares two candidates and executes eight actions from each selected 16-step chunk. Each method is evaluated over 50 trials per task.

\paragraphc{Results.}
As reported in Table~\ref{tab:realworld}, HDH improves success across all three real-world tasks, raising average success from 44.7\% to 67.3\%. The largest gain occurs on plushie placement, where success increases from 64\% to 94\%. In the inspected rollouts where the plushie starts farther from the gripper, the baseline sometimes closes the gripper before reaching the object, whereas HDH continues the approach before closing. These examples suggest that candidate selection can avoid premature grasp attempts without changing the underlying VLA. Gains also extend to the drawer and closet tasks, showing that action verification can improve task completion while keeping the VLA frozen.

%% file: tables/inference_time.tex
\begin{table}[t]
\centering
\caption{\textbf{Inference Time.}
End-to-end latency per policy query. Overhead is computed from
the median latency relative to the baseline.}
\label{tab:inference_time}

{\fontfamily{ptm}\selectfont
\small
\renewcommand{\arraystretch}{1.12}
\setlength{\tabcolsep}{6pt}

\begin{tabular}{@{}lrrr@{}}
\toprule
Method & p50 (ms) & p95 (ms) & Overhead vs.\ Baseline \\
\midrule
Baseline   & 681    & 760    & -- \\
Random     & 692    & 776    & +1.6\% \\
GPT-5      & 27,674 & 46,911 & +3963.7\% \\
GPT-6      & 7,403  & 10,270 & +987.1\% \\
HDM        & 716    & 799    & +5.1\% \\
HDH (Ours) & 718    & 796    & \textbf{+5.4\%} \\
\bottomrule
\end{tabular}
}
\end{table}

%% file: tables/realworld.tex
\begin{table}[t]
\centering
\caption{\textbf{Real-world task success.}
Results are evaluated over 50 trials per task and method.}
\label{tab:realworld}
\small
\setlength{\tabcolsep}{5pt}
\renewcommand{\arraystretch}{1.15}

\begin{tabularx}{\columnwidth}{
    @{}>{\raggedright\arraybackslash}X
    >{\centering\arraybackslash}p{0.17\columnwidth}
    >{\centering\arraybackslash}p{0.19\columnwidth}@{}
}
\toprule
Task & Baseline & \method{} (Ours) \\
\midrule
Plushie onto plate
& 64\% & \textbf{94\%} \\

Cube out of drawer; close drawer
& 38\% & \textbf{52\%} \\

Strawberry into closet; close door
& 32\% & \textbf{56\%} \\
\midrule
Average
& 44.7\% & \textbf{67.3\%} \\
\bottomrule
\end{tabularx}
\end{table}

%% file: sections/appendix.tex
\appendices

\section{Base Policy Implementation Details}
\label{app:policy}

Both base policies are \pifive{}~\cite{pi05} models fine-tuned with LoRA~\cite{lora} in openpi. After fine-tuning, the policy is frozen for verifier data collection, verifier training, and all evaluations. Table~\ref{tab:policy_settings} lists the settings for both domains.

\paragraphc{Simulation policy.}
The LIBERO policy is initialized from the \pifive{} base checkpoint. It receives the agent-view and wrist RGB images, an 8-D proprioceptive state, and the task instruction. The state contains the end-effector position (3-D), its axis-angle orientation (3-D), and the two gripper joint positions. LIBERO renders both views at $256\times256$. Following openpi's LIBERO preprocessing, we rotate each image by $180^\circ$ and resize it with padding to $224\times224$. The policy predicts chunks of $L=10$ 7-D actions, consisting of end-effector translation, rotation, and gripper commands.

The fine-tuning set contains ten public LIBERO demonstrations from each of five LIBERO-10 tasks (Table~\ref{tab:finetune_tasks}), 50 demonstrations in total. Four of these tasks are the evaluation tasks, and the fifth is used only for fine-tuning. Demonstrations are selected by fixed episode index and verified against per-episode checksums. Policy fine-tuning, verifier training, and evaluation use disjoint sets of initial states.

\paragraphc{Real-world policy.}
The DROID policy is initialized from the public \pifive{}-DROID checkpoint~\cite{droid}. It receives one external camera view and the wrist camera view, both resized with padding to $224\times224$. It also receives the seven joint positions, the gripper position, and the task instruction. It predicts chunks of $L=16$ 8-D actions, consisting of seven joint velocities and one gripper position. The fine-tuning set contains 80 teleoperated demonstrations across four tasks. State and action normalization statistics are inherited from the \pifive{}-DROID checkpoint.

\paragraphc{Fine-tuning.}
Both policies are trained with the conditional flow-matching objective of \pifive{}. LoRA adapters are added to the attention and feed-forward layers of the Gemma-2B backbone (rank 16) and the 300M action expert (rank 32). All other Gemma and action-expert weights are frozen, while the SigLIP image encoder~\cite{siglip} and the action projections remain trainable. States and actions are normalized to $[-1,1]$ using their 1st and 99th percentiles. For simulation, these statistics are computed from the fine-tuning set. We train for 5,000 steps and use the final checkpoint without exponential moving averaging or validation-based checkpoint selection.

\paragraphc{Inference.}
Each action chunk is generated from standard Gaussian noise with 10 forward Euler steps of the learned flow. The robot executes only the first $H=8$ actions before the next query: 8 of 10 actions in simulation and 8 of 16 on the real robot. With $K=2$, each candidate uses its own seeded noise. The first candidate's noise equals the sample that the $K=1$ baseline draws, so on identical observations every method sees the same first candidate.

\begin{table}[t]
\centering
\caption{\textbf{Base policy fine-tuning settings.}}
\label{tab:policy_settings}
\footnotesize
\renewcommand{\arraystretch}{1.1}
\setlength{\tabcolsep}{4pt}
\begin{tabular}{@{}lll@{}}
\toprule
Setting & Simulation & Real world \\
\midrule
Initialization & \pifive{} base & \pifive{}-DROID \\
Demonstrations & 50 (5 tasks) & 80 (4 tasks) \\
Camera views & 2 $\times$ $224^2$ & 2 $\times$ $224^2$ \\
State / action dim. & 8 / 7 & 8 / 8 \\
Chunk length $L$ & 10 & 16 \\
Executed prefix $H$ & 8 & 8 \\
\midrule
LoRA rank (backbone / expert) & \multicolumn{2}{l}{16 / 32} \\
Optimizer & \multicolumn{2}{l}{AdamW, $(\beta_1,\beta_2)=(0.9,0.95)$} \\
Weight decay & \multicolumn{2}{l}{$10^{-10}$} \\
Learning-rate schedule & \multicolumn{2}{l}{1,000 warmup steps, cosine decay} \\
Peak / final learning rate & \multicolumn{2}{l}{$2.5\times10^{-5}$ / $2.5\times10^{-6}$} \\
Gradient clipping & \multicolumn{2}{l}{Maximum norm 1.0} \\
Training steps & \multicolumn{2}{l}{5,000} \\
Global batch size & \multicolumn{2}{l}{32} \\
EMA & \multicolumn{2}{l}{None} \\
Precision & \multicolumn{2}{l}{bfloat16} \\
Training seed & \multicolumn{2}{l}{42} \\
Flow integration steps & \multicolumn{2}{l}{10 (Euler)} \\
\bottomrule
\end{tabular}
\end{table}

\begin{table}[t]
\centering
\caption{\textbf{LIBERO-10 fine-tuning tasks.} Ten demonstrations per task. The last column marks the evaluation tasks.}
\label{tab:finetune_tasks}
\small
\renewcommand{\arraystretch}{1.1}
\begin{tabularx}{\columnwidth}{@{}c>{\raggedright\arraybackslash}Xc@{}}
\toprule
ID & Instruction & Eval. \\
\midrule
0 & put both the alphabet soup and the tomato sauce in the basket & \checkmark \\
2 & turn on the stove and put the moka pot on it & \\
3 & put the black bowl in the bottom drawer of the cabinet and close it & \checkmark \\
6 & put the white mug on the plate and put the chocolate pudding to the right of the plate & \checkmark \\
7 & put both the alphabet soup and the cream cheese box in the basket & \checkmark \\
\bottomrule
\end{tabularx}
\end{table}

\section{Verifier Implementation Details}
\label{app:verifier}

\paragraphc{Inputs.}
At each query, we take the image embeddings of the external and wrist views from the frozen policy's prefix, which are the SigLIP outputs projected to the 2048-D Gemma width. We mean-pool each view's $16\times16$ grid to $8\times8$ with $2\times2$ windows. These tokens come from a separate prefix forward pass that does not change the policy's sampling noise, and they are stored in half precision. Proprioception and executed actions are normalized per dimension by mean and standard deviation, identically during training and inference. In simulation, these statistics come from the policy checkpoint. On the real robot, they are computed from the verifier's training rollouts.

\paragraphc{History token.}
HDH summarizes up to three preceding completed queries $j$. Each query is embedded as
\begin{equation}
e_j = W_v\,\bar{z}_j + W_s\,s_j + W_a\,\bar{U}_j,
\end{equation}
where $\bar{z}_j$ is the mean of the 128 visual tokens, $s_j$ is the normalized proprioception, and $\bar{U}_j$ is the mean of the $H$ executed actions. A single-layer GRU with hidden size 128 encodes the sequence $(e_j)$, and its last valid hidden state is appended to the observation context as one token. At the first query of an episode, when no completed query exists, a learned token $h_{\mathrm{none}}$ is used instead. At inference, the executed prefix of the selected candidate is added to the history after each query. Training and inference construct the history in the same way.

\paragraphc{Optimization.}
Each minibatch samples episodes uniformly and then one query uniformly within each sampled episode, so that long episodes, which are more often failures, do not dominate the loss. Queries with fewer than $H$ executed actions are discarded. We train with Adam for 40 epochs without weight decay, learning-rate schedule, or early stopping, and use the final checkpoint. Validation ROC AUC over held-out episodes is used only for monitoring. Table~\ref{tab:verifier_settings} lists the full configuration.

\begin{table}[t]
\centering
\caption{\textbf{Verifier settings.}}
\label{tab:verifier_settings}
\footnotesize
\renewcommand{\arraystretch}{1.1}
\setlength{\tabcolsep}{4pt}
\begin{tabular}{@{}lll@{}}
\toprule
Setting & Simulation & Real world \\
\midrule
Width / attention heads & \multicolumn{2}{l}{128 / 4} \\
Visual tokens & \multicolumn{2}{l}{$2\times64$ (2048-D)} \\
Context tokens (HDM / HDH) & \multicolumn{2}{l}{130 / 131} \\
Action prefix & $8\times7$ & $8\times8$ \\
Feed-forward network & \multicolumn{2}{l}{$128\rightarrow256\rightarrow128$} \\
Scoring head & \multicolumn{2}{l}{$128\rightarrow64\rightarrow1$} \\
History window / GRU width & \multicolumn{2}{l}{3 queries / 128} \\
Parameters (HDM / HDH) & \multicolumn{2}{l}{481K / 845K} \\
\midrule
Loss & \multicolumn{2}{l}{BCE on $1-y_e$} \\
Optimizer / learning rate & \multicolumn{2}{l}{Adam / $10^{-3}$} \\
Epochs & \multicolumn{2}{l}{40} \\
Batch size & 128 & 64 \\
Normalization statistics & Policy & Rollouts \\
Validation split & 20 of 100 episodes & 20\% of episodes \\
Training seeds per task & 3 & 1 \\
\midrule
Candidates $K$ & \multicolumn{2}{l}{2} \\
\bottomrule
\end{tabular}
\end{table}

\section{Simulation Evaluation Protocol}
\label{app:sim_protocol}

\paragraphc{Initial states.}
Verifier-training and evaluation initial states are generated by seeded environment resets from disjoint seed ranges. Training states that coincide with any of LIBERO's 50 canonical initial states are discarded. Each task uses 100 training states and 50 evaluation states, and all methods and runs share the same evaluation states.

\paragraphc{Episodes.}
Each episode begins with 10 idle steps with the gripper open, so that objects settle. Episodes end at task success, as determined by LIBERO's goal predicates, or after 520 control steps. Within a run, all methods share one policy server and derive their sampling noise from the task, initial state, and run index, so every method sees the same first candidate on identical observations. Random selection is seeded by the run index.

\paragraphc{Timing.}
We report end-to-end wall-clock time per query: episode duration divided by the number of policy queries. This includes policy sampling, verifier scoring, and simulator stepping on a single NVIDIA H100 80GB GPU.

\section{Real-World Evaluation Protocol}
\label{app:real_protocol}

\paragraphc{Control.}
The policy runs on a remote GPU server, and the robot workstation streams observations over a websocket. Joint-velocity and gripper commands are executed at 15~Hz. Each action is clipped to $[-1,1]$, and the gripper command is binarized at 0.5. To hide network latency, the next query is sent two control steps before the current prefix is exhausted. Episodes are limited to 600 control steps (40~s). Each instruction is routed to its task's verifier by keyword.

\paragraphc{Outcomes and verifier data.}
After each episode, the operator labels it as success, failure, intervention, or error. For verifier training, interventions count as failures and error episodes are discarded. For each retained query, we record the exact policy inputs and the sampled chunk.

\paragraphc{Statistical analysis.}
Table~\ref{tab:real_stats} reports two-sided Fisher's exact tests on per-task and pooled success counts, without multiple-comparison correction. The tests treat trials as independent. The improvement is significant at $\alpha=0.05$ for plushie placement, the closet task, and the pooled comparison, but not for the drawer task.

\begin{table}[t]
\centering
\caption{\textbf{Real-world significance tests.} Successes out of 50 trials per task and method, with two-sided Fisher's exact $p$-values.}
\label{tab:real_stats}
\footnotesize
\renewcommand{\arraystretch}{1.1}
\begin{tabular}{@{}lccc@{}}
\toprule
Task & Baseline & \method{} & $p$ \\
\midrule
Plushie onto plate & 32/50 & 47/50 & $3.9\times10^{-4}$ \\
Cube out of drawer & 19/50 & 26/50 & 0.23 \\
Strawberry into closet & 16/50 & 28/50 & 0.026 \\
\midrule
Pooled & 67/150 & 101/150 & $1.2\times10^{-4}$ \\
\bottomrule
\end{tabular}
\end{table}


